\documentclass[11pt,a4paper]{article}
\usepackage{times,latexsym}
\usepackage{url}
\usepackage[T1]{fontenc}

\usepackage[acceptedWithA]{tacl2018v2}

\usepackage{xspace,mfirstuc,tabulary, enumitem}

\newif\iftaclinstructions
\taclinstructionsfalse 
\iftaclinstructions
\renewcommand{\confidential}{}
\renewcommand{\anonsubtext}{(No author info supplied here, for consistency with
TACL-submission anonymization requirements)}
\newcommand{\instr}
\fi

\iftaclpubformat 

\else

\fi

\usepackage{array,multirow,graphicx,adjustbox}

\title{Word Acquisition in Neural Language Models}

\author{Tyler A. Chang$^{1,2}$, \quad Benjamin K. Bergen$^1$ \\
$^1$Department of Cognitive Science \\
$^2${Halıcıoğlu} Data Science Institute \\
University of California, San Diego \\
{\texttt{$\{$tachang, bkbergen$\}$@ucsd.edu}}
}

\date{}

\begin{document}
\maketitle
\begin{abstract}
We investigate how neural language models acquire individual words during training, extracting learning curves and ages of acquisition for over 600 words on the MacArthur-Bates Communicative Development Inventory \cite{fenson-etal-2007-macarthur}.
Drawing on studies of word acquisition in children, we evaluate multiple predictors for words' ages of acquisition in LSTMs, BERT, and GPT-2.
We find that the effects of concreteness, word length, and lexical class are pointedly different in children and language models, reinforcing the importance of interaction and sensorimotor experience in child language acquisition.
Language models rely far more on word frequency than children, but like children, they exhibit slower learning of words in longer utterances.
Interestingly, models follow consistent patterns during training for both unidirectional and bidirectional models, and for both LSTM and Transformer architectures.
Models predict based on unigram token frequencies early in training, before transitioning loosely to bigram probabilities, eventually converging on more nuanced predictions.
These results shed light on the role of distributional learning mechanisms in children, while also providing insights for more human-like language acquisition in language models.
\end{abstract}

\section{Introduction}
Language modeling, predicting words from context, has grown increasingly popular as a pre-training task in NLP in recent years; neural language models such as BERT \cite{devlin-etal-2019-bert}, ELMo \cite{peters-etal-2018-deep}, and GPT \cite{brown-etal-2020-language} have produced state-of-the-art performance on a wide range of NLP tasks.
There is now a substantial amount of work assessing the linguistic information encoded by language models \cite{rogers-etal-2020-primer}; in particular, behavioral approaches from psycholinguistics and cognitive science have been applied to study language model predictions (\citealp{futrell-etal-2019-neural}; \citealp{ettinger-2020-bert}).
From a cognitive perspective, language models are of theoretical interest as distributional models of language, agents that learn exclusively from statistics over language (\citealp{boleda-2020-distributional}; \citealp{lenci-2018-distributional}).

\begin{figure}
    \centering
    \adjustbox{trim=0cm 0cm 0cm 0cm}{%
    \includegraphics[width=7.5cm]{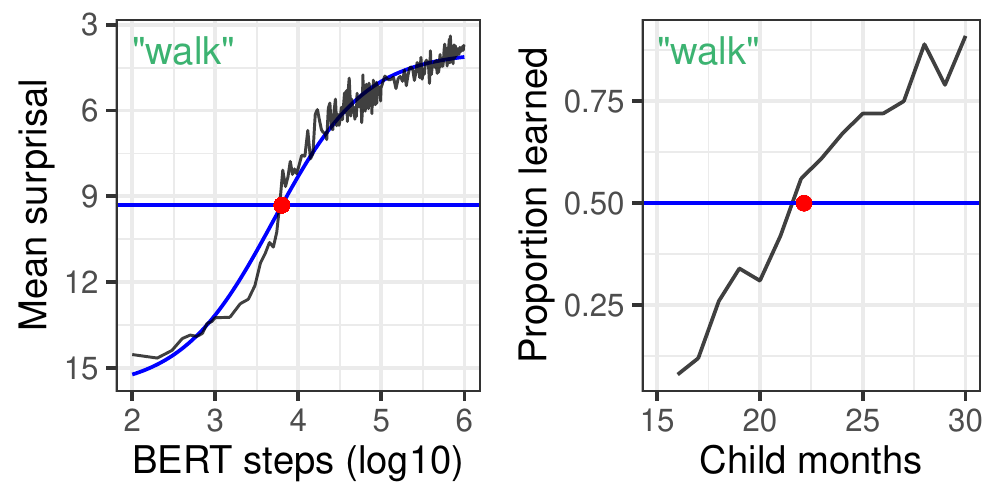}
    }
    \caption{Learning curves for the word ``walk'' in a BERT language model and human children. Blue horizontal lines indicate age of acquisition cutoffs.
    The blue curve represents the fitted sigmoid function based on the language model surprisals during training (black).
    Child data obtained from \citet{frank-etal-2017-wordbank}.}
    \label{fig:walk}
\end{figure}

However, previous psycholinguistic studies of language models have nearly always focused on fully-trained models, precluding comparisons to the wealth of literature on human language acquisition. There are limited exceptions.
\citet{rumelhart-mcclelland-1986-learning} famously studied past tense verb form learning in phoneme-level neural networks during training, a study which was replicated in more modern character-level recurrent neural networks \cite{kirov-cotterell-2018-recurrent}.
However, these studies focused only on sub-word features.
There remains a lack of research on language acquisition in contemporary language models, which encode higher level features such as syntax and semantics.

As an initial step towards bridging the gap between language acquisition and language modeling, we present an empirical study of word acquisition during training in contemporary language models, including LSTMs, GPT-2, and BERT.
We consider how variables such as word frequency, concreteness, and lexical class contribute to words' ages of acquisition in language models.
Each of our selected variables has effects on words' ages of acquisition in children; our language model results allow us to identify the extent to which each effect in children can or cannot be attributed in principle to distributional learning mechanisms.

Finally, to better understand how computational models acquire language, we identify consistent patterns in language model training across architectures.
Our results suggest that language models may acquire traditional distributional statistics such as unigram and bigram probabilities in a systematic way.
Understanding how language models acquire language can lead to better architectures and task designs for future models, while also providing insights into distributional learning mechanisms in people.

\section{Related work}
Our work draws on methodologies from word acquisition studies in children and psycholinguistic evaluations of language models.
In this section, we briefly outline both lines of research.

\subsection{Child word acquisition}
Child development researchers have previously studied word acquisition in children, identifying variables that help predict words' ages of acquisition in children.
In Wordbank, \citet{frank-etal-2017-wordbank} compiled reports from parents reporting when their child produced each word on the MacArthur-Bates Communicative Development Inventory (CDI; \citealp{fenson-etal-2007-macarthur}).
For each word $w$, \citet{braginsky-etal-2016-tomorrow} fitted a logistic curve predicting the proportion of children that produce $w$ at different ages; they defined a word's age of acquisition as the age at which 50\% of children produce $w$.
Variables such as word frequency, word length, lexical class, and concreteness were found to influence words' ages of acquisition in children across languages.
Recently, it was shown that fully-trained LSTM language model surprisals are also predictive of words' ages of acquisition in children \cite{portelance-etal-2020-predicting}.
However, no studies have evaluated ages of acquisition in language models themselves.

\subsection{Evaluating language models}
Recently, there has been substantial research evaluating language models using psycholinguistic approaches, reflecting a broader goal of interpreting language models (BERTology; \citealp{rogers-etal-2020-primer}).
For instance, \citet{ettinger-2020-bert} used the output token probabilities from BERT in carefully constructed sentences, finding that BERT learns commonsense and semantic relations to some degree, although it struggles with negation.
\citet{gulordava-etal-2018-colorless} found that LSTM language models recognize long distance syntactic dependencies; however, they still struggle with more complicated constructions \cite{marvin-linzen-2018-targeted}.

These psycholinguistic methodologies do not rely on specific language model architectures or fine-tuning on a probe task.
Notably, because these approaches rely only on output token probabilities from a given language model, they are well-suited to evaluations early in training, when fine-tuning on downstream tasks is unfruitful.
That said, previous language model evaluation studies have focused on fully-trained models, progressing largely independently from human language acquisition literature.
Our work seeks to bridge this gap.

\section{Method}
We trained unidirectional and bidirectional language models with LSTM and Transformer architectures.
We quantified each language model's age of acquisition for each word in the CDI \cite{fenson-etal-2007-macarthur}. Similar to word acquisition studies in children, we identified predictors for words' ages of acquisition in language models.\footnote{Code and data are available at \url{https://github.com/tylerachang/word-acquisition-language-models}.}

\subsection{Language models}
\paragraph{Datasets and training}
Language models were trained on a combined corpus containing the BookCorpus \citep{zhu-etal-2015-aligning} and WikiText-103 datasets \citep{merity-etal-2017-pointer}.
Following \citet{devlin-etal-2019-bert}, each input sequence was a sentence pair; the training dataset consisted of 25.6M sentence pairs.
The remaining sentences (5.8M pairs) were used for evaluation and to generate word learning curves.
Sentences were tokenized using the unigram language model tokenizer implemented in SentencePiece \cite{kudo-richardson-2018-sentencepiece}.
Models were trained for 1M steps, with batch size 128 and learning rate 0.0001.
As a metric for overall language model performance, we report evaluation perplexity scores in Table \ref{tab:perplexity}.
We include evaluation loss curves, full training details, and hyperparameters in Appendix \ref{app:training-details}.

\begin{table}[t]
    \centering
    \small
    \renewcommand{\arraystretch}{1.1}
    \begin{tabular}{|r|c|c|}
        \cline{2-3}
         \multicolumn{1}{c|}{} & \# Parameters & Perplexity \\
        \hline
        LSTM & 37M & 54.8 \\
        GPT-2 & 108M & 30.2 \\
        BiLSTM & 51M & 9.0 \\
        BERT & 109M & 7.2 \\
        \hline
    \end{tabular}
    \caption{Parameter counts and evaluation perplexities for the trained language models.
    For reference, the pre-trained BERT base model from Huggingface reached a perplexity of 9.4 on our evaluation set.
    Additional perplexity comparisons with comparable models are included in Appendix \ref{app:training-details}.
    }
    \label{tab:perplexity}
\end{table}

\begin{figure*}
    \centering
    \adjustbox{trim=0cm 0cm 0cm 0cm}{%
    \includegraphics[width=3.6cm]{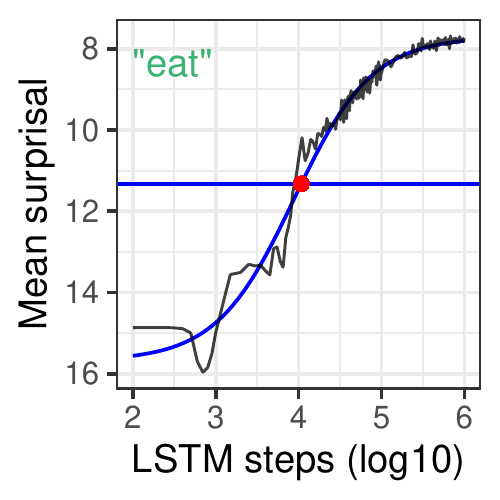}
    \includegraphics[width=3.6cm]{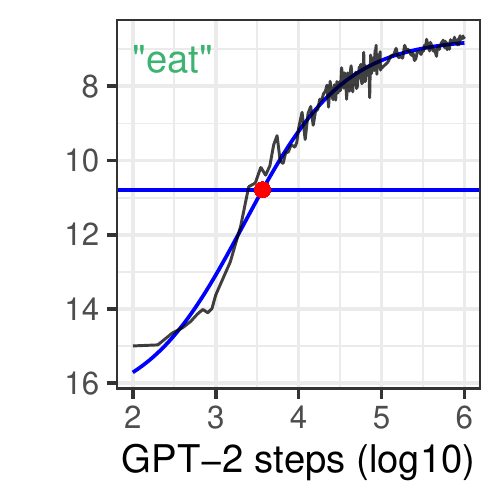}
    \includegraphics[width=3.6cm]{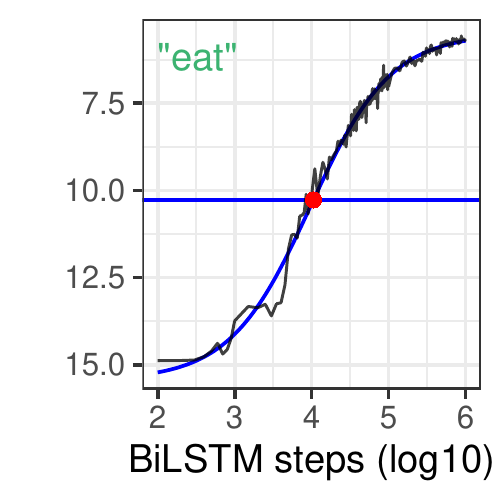}
    \includegraphics[width=3.6cm]{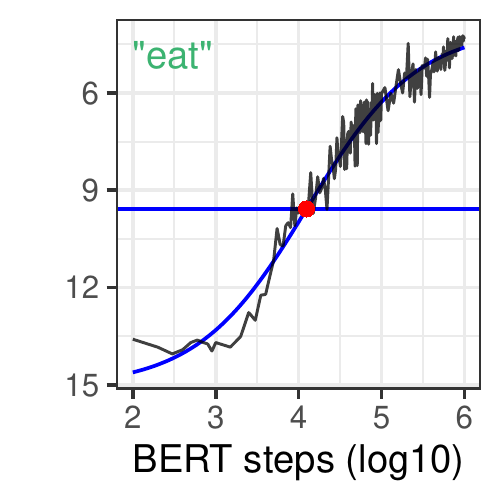}
    }
    \caption{Learning curves for the word ``eat'' for all four language model architectures. Blue horizontal lines indicate age of acquisition cutoffs, and blue curves represent fitted sigmoid functions.}
    \label{fig:eat}
\end{figure*}

\paragraph{Transformers}
The two Transformer models followed the designs of GPT-2 \cite{radford-2019-language} and BERT \cite{devlin-etal-2019-bert}, allowing us to evaluate both a unidirectional and bidirectional Transformer language model.
GPT-2 was trained with the causal language modeling objective, where each token representation is used to predict the next token; the masked self-attention mechanism allows tokens to attend only to previous tokens in the input sequence.
In contrast, BERT used the masked language modeling objective, where masked tokens are predicted from surrounding tokens in both directions.

Our BERT model used the base size model from \citet{devlin-etal-2019-bert}.
Our GPT-2 model used the similar-sized model from \citet{radford-2019-language}, equal in size to the original GPT model.
Parameter counts are listed in Table \ref{tab:perplexity}.
Transformer models were trained using the Huggingface Transformers library \cite{wolf-etal-2020-transformers}.

\paragraph{LSTMs}
We also trained both a unidirectional and bidirectional LSTM language model, each with three stacked LSTM layers.
Similar to GPT-2, the unidirectional LSTM predicted the token at time $t$ from the hidden state at time $t-1$.
The bidirectional LSTM (BiLSTM) predicted the token at time $t$ from the sum of the hidden states at times $t-1$ and $t+1$ \cite{aina-etal-2019-putting}.

\subsection{Learning curves and ages of acquisition}
We sought to quantify each language model's ability to predict individual words over the course of training.
We considered all words in the CDI that were considered one token by the language models (611 out of 651 words).

For each such token $w$, we identified up to 512 occurrences of $w$ in the held-out portion of the language modeling dataset.\footnote{We only selected sentence pairs with at least eight tokens of context, unidirectionally or bidirectionally depending on model architecture.
Thus, the unidirectional and bidirectional samples differed slightly.
Most tokens (92.3\%) had the maximum of 512 samples both unidirectionally and bidirectionally, and all tokens had at least 100 samples in both cases.}
To evaluate a language model at training step $s$, we fed each sentence pair into the model, attempting to predict the masked token $w$.
We computed the surprisal: $-\log_2(P(w))$ averaged over all occurrences of $w$ to quantify the quality of the models' predictions for word $w$ at step $s$ (\citealp{levy-2008-expectation}; \citealp{goodkind-bicknell-2018-predictive}).

We computed this average surprisal for each target word at approximately 200 different steps during language model training, sampling more heavily from earlier training steps, prior to model convergence.
The selected steps are listed in Appendix \ref{app:training-details}.
By plotting surprisals over the course of training, we obtained a learning curve for each word, generally moving from high surprisal to low surprisal.
The surprisal axis in our plots is reversed to reflect increased understanding over the course of training, consistent with plots showing increased proportions of children producing a given word over time \cite{frank-etal-2017-wordbank}.

For each learning curve (4 language model architectures $\times$ 611 words), we fitted a sigmoid function to model the smoothed acquisition of word $w$.
Sample learning curves are shown in Figure \ref{fig:walk} and Figure \ref{fig:eat}.

\paragraph{Age of acquisition}
To extract age of acquisition from a learning curve, we established a cutoff surprisal where we considered a given word ``learned.''
In child word acquisition studies, an analogous cutoff is established when 50\% of children produce a word \cite{braginsky-etal-2016-tomorrow}.

Following this precedent, we determined our cutoff to be 50\% between a baseline surprisal (predicting words based on random chance) and the minimum surprisal attained by the model for word $w$.
We selected the random chance baseline to best reflect a language model's ability to predict a word with no access to any training data, similar to an infant's language-specific knowledge prior to any linguistic exposure.
We selected minimum surprisal as our other bound to reflect how well a particular word can eventually be learned by a particular language model, analogous to an adult's understanding of a given word.

For each learning curve, we found the intersection between the fitted sigmoid and the cutoff surprisal value.
We defined age of acquisition for a language model as the corresponding training step, on a log10 scale.
Sample cutoffs and ages of acquisition are shown in Figure \ref{fig:walk} and Figure \ref{fig:eat}.

\subsection{Predictors for age of acquisition}
As potential predictors for words' ages of acquisition in language models, we selected variables that are predictive of age of acquisition in children \cite{braginsky-etal-2016-tomorrow}.
When predicting ages of acquisition in language models, we computed word frequencies and utterance lengths over the language model training corpus.
Our five selected predictors were:
\begin{itemize}
    \setlength\itemsep{0.1em}
    \item Log-frequency: the natural log of the word's per-1000 token frequency.
    \item MLU: we computed the mean length of utterance as the mean length of sequences containing a given word.\footnote{We also considered a unidirectional MLU metric (counting only previous tokens) for the unidirectional models, finding that it produced similar results.}
    MLU has been used as a metric for the complexity of syntactic contexts in which a word appears \cite{roy-etal-2015-predicting}.
    \item n-chars: as in \citet{braginsky-etal-2016-tomorrow}, we used the number of characters in a word as a coarse proxy for the length of a word.
    \item Concreteness: we used human-generated concreteness norms from \citet{brysbaert-etal-2014-concreteness}, rated on a five-point scale.
    We imputed missing values (3\% of words) using the mean concreteness score.
    \item Lexical class: we used the lexical classes annotated in Wordbank.
    Possible lexical classes were Noun, Verb, Adjective, Function Word, and Other.
\end{itemize}
We ran linear regressions with linear terms for each predictor.
To determine statistical significance for each predictor, we ran likelihood ratio tests, comparing the overall regression (including the target predictor) with a regression including all predictors except the target.
To determine the direction of effect for each continuous predictor, we used the sign of the coefficient in the overall regression.

\begin{table*}[t]
    \centering
    \begin{tabular}{|r|c|c|c|c|c|}
        \cline{2-6}
         \multicolumn{1}{c|}{} & LSTM & GPT-2 & BiLSTM & BERT & Children \\
        \hline
        Log-frequency & $^{***} \hspace{0.1cm} (-)$ & $^{***} \hspace{0.1cm} (-)$ & $^{***} \hspace{0.1cm} (-)$ & $^{***} \hspace{0.1cm} (-)$ & $^{***} \hspace{0.1cm} (-)$ \\
        MLU & & $\phantom{^{*}}^{**} \hspace{0.1cm} (+)$ & $^{***} \hspace{0.1cm} (+)$ & $^{***} \hspace{0.1cm} (+)$ & $^{***} \hspace{0.1cm} (+)$ \\
        n-chars & $^{***} \hspace{0.1cm} (-)$ & $^{***} \hspace{0.1cm} (-)$ & $^{***} \hspace{0.1cm} (-)$ & $^{***} \hspace{0.1cm} (-)$ & $\phantom{^{*}}^{**} \hspace{0.1cm} (+)$ \\
        Concreteness & & & & & $^{***} \hspace{0.1cm} (-)$ \\
        Lexical class & $^{***} \hspace{0.1cm} \phantom{(+)}$ & $^{***} \hspace{0.1cm} \phantom{(+)}$ & & & $^{***} \hspace{0.1cm} \phantom{(+)}$ \\
        \hline
        $R^2$ & 0.93 & 0.92 & 0.95 & 0.94 & 0.43 \\
        \hline
    \end{tabular}
    \caption{Significant predictors for a word's age of acquisition are marked by asterisks ($p < 0.05^{*}$; $p < 0.01^{**}$; $p < 0.001^{***}$).
    Signs of coefficients are notated in parentheses. The $R^2$ denotes the adjusted $R^2$ in a regression using all five predictors.}
    \label{tab:regression}
\end{table*}

As a potential concern for interpreting regression coefficient signs, we assessed collinearities between predictors by computing the variance inflation factor (VIF) for each predictor.
No VIF exceeded $5.0$,\footnote{Common VIF cutoff values are $5.0$ and $10.0$.} although we did observe moderate correlations between log-frequency and n-chars ($r = -0.49$), and between log-frequency and concreteness ($r = -0.64$).
These correlations are consistent with those identified for child-directed speech in \citet{braginsky-etal-2016-tomorrow}.
To ease collinearity concerns, we considered single-predictor regressions for each predictor, using adjusted predictor values after accounting for log-frequency (residuals after regressing the predictor over log-frequency).
In all cases, the coefficient sign in the adjusted single predictor regression was consistent with the sign of the coefficient in the overall regression.

When lexical class (the sole categorical predictor) reached significance based on the likelihood ratio test, we ran a one-way analysis of covariance (ANCOVA) with log-frequency as a covariate.
The ANCOVA ran a standard ANOVA on the age of acquisition residuals after regressing over log-frequency.
We used Tukey's honestly significant difference (HSD) test to assess pairwise differences between lexical classes.

\subsection{Age of acquisition in children}
\label{sec:aoa-in-children}
For comparison, we used the same variables to predict words' ages of acquisition in children, as in \citet{braginsky-etal-2016-tomorrow}.
We obtained smoothed ages of acquisition for children from the Wordbank dataset \cite{frank-etal-2017-wordbank}.
When predicting ages of acquisition in children, we computed word frequencies and utterance lengths over the North American English CHILDES corpus of child-directed speech \cite{macwhinney-2000-childes}.

Notably, CHILDES contained much shorter sentences on average than the language model training corpus (mean sentence length $4.50$ tokens compared to $15.14$ tokens).
CDI word log-frequencies were only moderately correlated between the two corpora ($r=0.78$).
This aligns with previous work finding that child-directed speech contains on average fewer words per utterance, smaller vocabularies, and simpler syntactic structures than adult-directed speech \cite{soderstrom-2007-reevaluating}.
These differences were likely compounded by differences between spoken language in the CHILDES corpus and written language in the language model corpus.
We computed word frequencies and MLUs separately over the two corpora to ensure that our predictors accurately reflected the learning environments of children and the language models.

We also note that the language model training corpus was much larger overall than the CHILDES corpus.
CHILDES contained 7.5M tokens, while the language model corpus contained 852.1M tokens.
Children are estimated to hear approximately 13K words per day \cite{gilkerson-etal-2017-mapping}, for a total of roughly 19.0M words during their first four years of life.
Because contemporary language models require much more data than children hear, the models do not necessarily reflect how children would learn if restricted solely to linguistic input.
Instead, the models serve as examples of relatively successful distributional learners, establishing how one might expect word acquisition to progress according to effective distributional mechanisms.

\begin{figure}
    \centering
    \adjustbox{trim=0cm 0cm 0cm 0cm}{%
    \includegraphics[width=3.5cm]{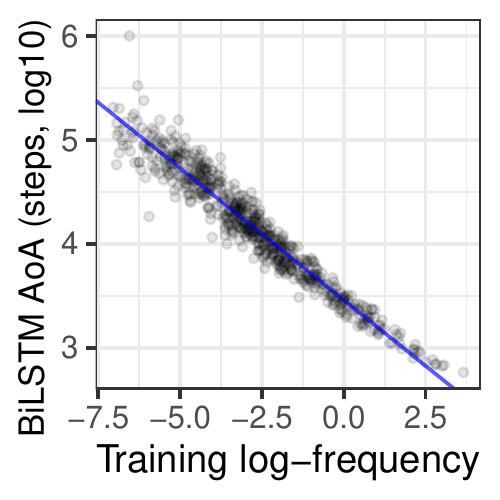}
    \includegraphics[width=3.5cm]{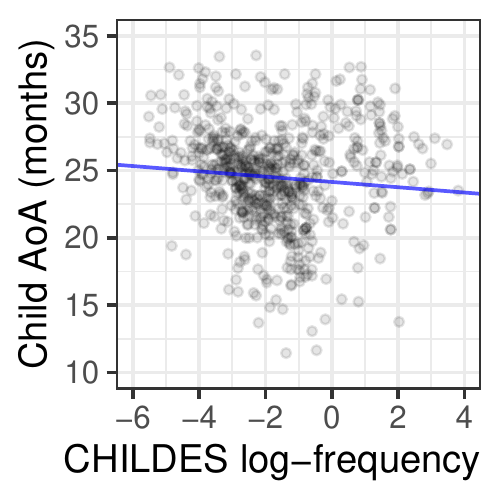}
    }
    \caption{Effects of log-frequency on words' ages of acquisition (AoA) in the BiLSTM and children. The BiLSTM was the language model architecture with the largest effect of log-frequency (adjusted $R^2 = 0.94$).}
    \label{fig:frequency}
\end{figure}

\section{Results}
Significant predictors of age of acquisition are shown in Table \ref{tab:regression}, comparing children and each of the four language model architectures.

\paragraph{Log-frequency}
In children and all four language models, more frequent words were learned earlier (a negative effect on age of acquisition).
As shown in Figure \ref{fig:frequency}, this effect was much more pronounced in language models (adjusted $R^2 = 0.91$ to $0.94$) than in children (adjusted $R^2 = 0.01$).\footnote{Because function words are frequent but acquired later by children, a quadratic model of log-frequency on age of acquisition in children provided a slightly better fit ($R^2 = 0.03$) if not accounting for lexical class. A quadratic model of log-frequency also provided a slightly better fit for unidirectional language models ($R^2 = 0.93$ to $0.94$), particularly for high-frequency words; in language models, this could be due either to a floor effect on age of acquisition for high-frequency words or to slower learning of function words. Regardless, significant effects of other predictors remained the same when using a quadratic model for log-frequency.}
The sizeable difference in log-frequency predictivity emphasizes the fact that language models learn exclusively from distributional statistics over words, while children have access to additional social and sensorimotor cues.

\paragraph{MLU}
Except in unidirectional LSTMs, MLU had a positive effect on a word's age of acquisition in language models.
Interestingly, we might have expected the opposite effect (particularly in Transformers) if additional context (longer utterances) facilitated word learning.
Instead, our results are consistent with effects of MLU in children; words in longer utterances are learned later, even after accounting for other variables.
The lack of effect in unidirectional LSTMs could simply be due to LSTMs being the least sensitive to contextual information of the models under consideration. The positive effect of MLU in other models suggests that complex syntactic contexts may be more difficult to learn through distributional learning alone, which might partly explain why children learn words in longer utterances more slowly.

\paragraph{n-chars}
There was a negative effect of n-chars on age of acquisition in all four language models; longer words were learned earlier. 
This contrasts with children, who acquire shorter words earlier.
This result is particularly interesting because the language models we used have no information about word length.
We hypothesize that the effect of n-chars in language models may be driven by polysemy, which is not accounted for in our regressions.
Shorter words tend to be more polysemous (a greater diversity of meanings; \citealp{casas-etal-2019-polysemy}), which could lead to slower learning in language models.
In children, this effect may be overpowered by the fact that shorter words are easier to parse and produce.

\paragraph{Concreteness}
Although children overall learn more concrete words earlier, the language models showed no significant effects of concreteness on age of acquisition.
This entails that the effects in children cannot be explained by correlations between concrete words and easier distributional learning contexts.
Again, this highlights the importance of sensorimotor experience and conceptual development in explaining the course of child language acquisition.

\paragraph{Lexical class}
The bidirectional language models showed no significant effects of lexical class on age of acquisition.
In other words, the differences between lexical classes were sufficiently accounted for by the other predictors for BERT and the BiLSTM.
However, in the unidirectional language models (GPT-2 and the LSTM), nouns and function words were acquired later than adjectives and verbs.\footnote{Significant pairwise comparisons between lexical classes are listed in Appendix \ref{app:pairwise}.}
This contrasts with children learning English, who on average acquired nouns earlier than adjectives and verbs, acquiring function words last.\footnote{There is ongoing debate around the existence of a universal ``noun bias'' in early word acquisition.
For instance, Korean and Mandarin-speaking children have been found to acquire verbs earlier than nouns, although this effect appears sensitive to context and the measure of vocabulary acquisition (\citealp{choi-gopnik-1995-early}; \citealp{tardif-etal-1999-putting}).}

Thus, children's early acquisition of nouns cannot be explained by distributional properties of English nouns, which are acquired later by unidirectional language models.
This result is compatible with the hypothesis that nouns are acquired earlier because they often map to real world objects; function words might be acquired later because their meanings are less grounded in sensorimotor experience.
It has also been argued that children might have an innate bias to learn objects earlier than relations and traits \cite{markman-1994-constraints}.
Lastly, it is possible that the increased salience of sentence-final positions (which are more likely to contain nouns in English and related languages) facilitates early acquisition of nouns in children  \cite{caselli-etal-1995-cross}.
Consistent with these hypotheses, our results suggest that English verbs and adjectives may be easier to learn from a purely distributional perspective, but children acquire nouns earlier based on sensorimotor, social, or cognitive factors.

\begin{table}[t]
    \centering
    \small
    \renewcommand{\arraystretch}{1.1}
    \begin{tabular}{|>{\raggedleft}p{0.6cm}|>{\raggedright}p{2.7cm}|>{\raggedright}p{2.7cm}|}
        \cline{2-3}
        \multicolumn{1}{c|}{} & Language models & Children \tabularnewline
        \hline
        First & \footnotesize \textit{a, and, for, he, her, his, I, it, my, of, on, she, that, the, to, was, with, you} \normalsize & \footnotesize \textit{baby, ball, bye, daddy, dog, hi, mommy, moo, no, shoe, uh, woof, yum} \normalsize \tabularnewline
        \hline
        Last & \footnotesize \textit{bee, bib, choo, cracker, crayon, giraffe, glue, kitty, moose, pancake, popsicle, quack, rooster, slipper, tuna, yum, zebra} \normalsize & \footnotesize \textit{above, basement, beside, country, downtown, each, hate, if, poor, walker, which, would, yourself} \normalsize \tabularnewline
        \hline
    \end{tabular}
    \caption{First and last words acquired by the language models and children.
    For language models, we identified words that were in the top or bottom 5\% of ages of acquisition for all models.
    For children, we identified words in the top or bottom 2\% of ages of acquisition.}
    \label{tab:first-last-words}
\end{table}

\subsection{First and last learned words}
As a qualitative analysis, we compared the first and last words acquired by the language models and children, as shown in Table \ref{tab:first-last-words}.
In line with our previous results, the first and last words learned by the language models were largely determined by word frequencies.
The first words acquired by the models were all in the top 3\% of frequent words, and the last acquired words were all in the bottom 15\%.
Driven by this effect, many of the first words learned by the language models were function words or pronouns.
In contrast, many of the first words produced by children were single-word expressions, such as greetings, exclamations, and sounds.
Children acquired several highly frequent words late, such as ``if,'' which is in the 90th frequency percentile of the CHILDES corpus.
Of course, direct comparisons between the first and last words acquired by the children and language models are confounded by differing datasets and learning environments, as detailed in Section \ref{sec:aoa-in-children}.

\subsection{Age of acquisition vs. minimum surprisal}
Next, we assessed whether a word's age of acquisition in a language model could be predicted from how well that word was learned in the fully-trained model.
To do this, we considered the minimum surprisal attained by each language model for each word.
We found a significant effect of minimum surprisal on age of acquisition in all four language models, even after accounting for all five other predictors (using likelihood ratio tests; $p < 0.001$).
In part, this is likely because the acquisition cutoff for each word's fitted sigmoid was dependent on the word's minimum surprisal.

It could then be tempting to treat minimum surprisal as a substitute for age of acquisition in language models; this approach would require only publicly-available fully-trained language models.
Indeed, the correlation between minimum surprisal and age of acquisition was substantial (Pearson's $r = 0.88$ to $0.92$).
However, this correlation was driven largely by effects of log-frequency, which had a large negative effect on both metrics.
When adjusting minimum surprisal and age of acquisition for log-frequency (using residuals after linear regressions), the correlation decreased dramatically (Pearson's $r = 0.22$ to $0.46$).
While minimum surprisal accounts for a significant amount of variance in words' ages of acquisition, the two metrics are not interchangeable.

\subsection{Alternative age of acquisition definitions}
\begin{figure*}
    \centering
    \adjustbox{trim=0cm 0cm 0cm 0cm}{%
    \includegraphics[width=3.6cm]{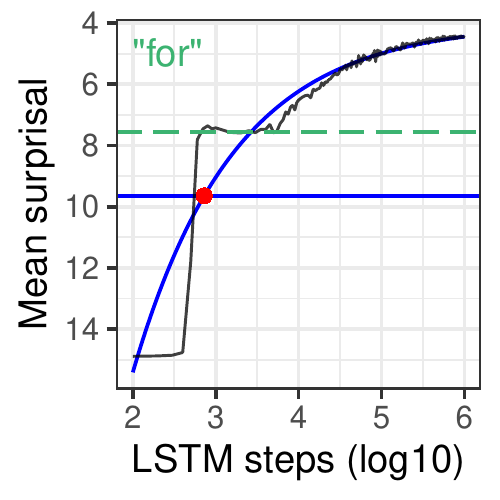}
    \includegraphics[width=3.6cm]{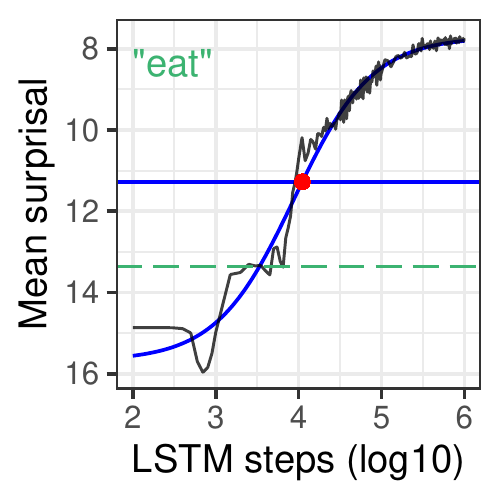}
    \includegraphics[width=3.6cm]{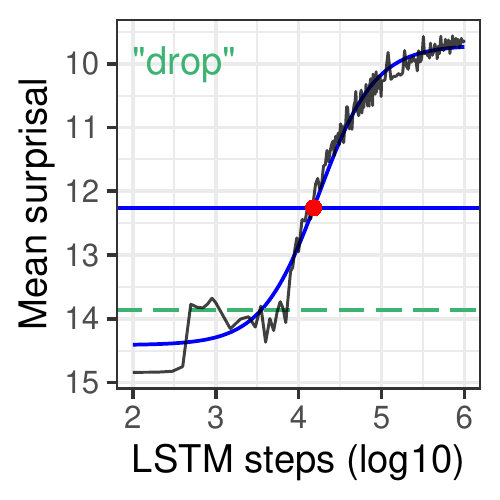}
    \includegraphics[width=3.6cm]{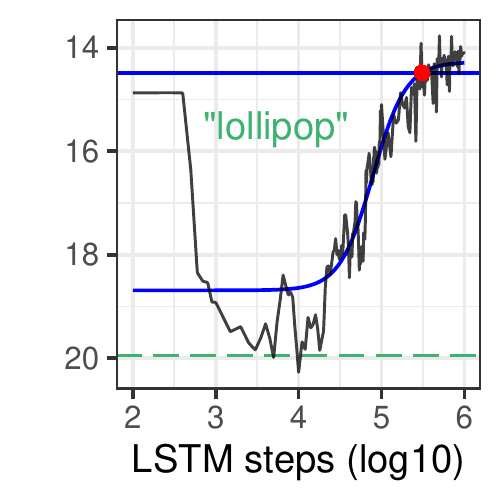}
    }
    \caption{LSTM learning curves for the words ``for,'' ``eat,'' ``drop,'' and ``lollipop.'' Blue horizontal lines indicate age of acquisition cutoffs, and blue curves represent fitted sigmoid functions. Green dashed lines indicate the surprisal if predicting solely based on unigram probabilities (raw token frequencies). Early in training, language model surprisals tended to shift towards the unigram frequency-based surprisals.}
    \label{fig:freq-curves}
\end{figure*}

Finally, we considered alternative operationalizations of words' ages of acquisition in language models.
For instance, instead of defining an acquisition cutoff at 50\% between random chance and the minimum surprisal for each word, we could consider the midpoint of each fitted sigmoid curve.
This method would be equivalent to defining upper and lower surprisal baselines at the upper and lower asymptotes of the fitted sigmoid, relying on the assumption that these asymptotes roughly approximate surprisal values before and after training.
However, this assumption fails in cases where the majority of a word's learning curve is modeled by only a sub-portion of the fitted sigmoid.
For example, for the word ``for'' in Figure \ref{fig:freq-curves}, the high surprisal asymptote is at $156753.5$, compared to a random chance surprisal of $14.9$ and a minimum surprisal of $4.4$.
Using the midpoint age of acquisition in this case would result in an age of acquisition of $-9.6$ steps (log10).

We also considered alternative cutoff proportions (replacing 50\%) in our original age of acquisition definition.
We considered cutoffs at each possible increment of 10\%.
The signs of nearly all significant coefficients in the overall regressions (see Table \ref{tab:regression}) remained the same for all language models regardless of cutoff proportion.\footnote{The only exception was a non-significant positive coefficient for n-chars in BERT with a 90\% acquisition cutoff.}

\section{Language model learning curves}
The previous sections identified factors that predict words' ages of acquisition in language models.
We now proceed with a qualitative analysis of the learning curves themselves.
We found that language models learn traditional distributional statistics in a systematic way.

\subsection{Unigram probabilities}
First, we observed a common pattern in word learning curves across model architectures.
As expected, each curve began at the surprisal value corresponding to random chance predictions.
Then, as shown in Figure \ref{fig:freq-curves}, many curves shifted towards the surprisal value corresponding to raw unigram probabilities (i.e. based on raw token frequencies).
This pattern was particularly pronounced in LSTM-based language models, although it appeared in all architectures.
Interestingly, the shift occurred even if the unigram surprisal was higher (or ``worse'') than random-chance surprisal, as demonstrated by the word ``lollipop'' in Figure \ref{fig:freq-curves}.
Thus, we posited that language models pass through an early stage of training where they approximate unigram probabilities.

To test this hypothesis, we aggregated each model's predictions for randomly masked tokens in the evaluation dataset (16K sequences), including tokens not on the CDI.
For each saved training step, we computed the average Kullback-Leibler (KL) divergence between the model predictions and the unigram frequency distribution.
For comparison, we also computed the KL divergence with a uniform distribution (random chance) and with the one-hot true token distribution.
We note that the KL divergence with the one-hot true token distribution is equivalent to the cross-entropy loss function using log base two.\footnote{All KL divergences were computed using log base two. KL divergences were computed as $\textrm{KL}(\textbf{y}_{ref}, \hat{\textbf{y}})$, where $\hat{\textbf{y}}$ was the model's predicted probability distribution and $\textbf{y}_{ref}$ was the reference distribution.}

\begin{figure*}
    \phantom{\_}\hspace{0.5cm}
    \includegraphics[width=5.7cm]{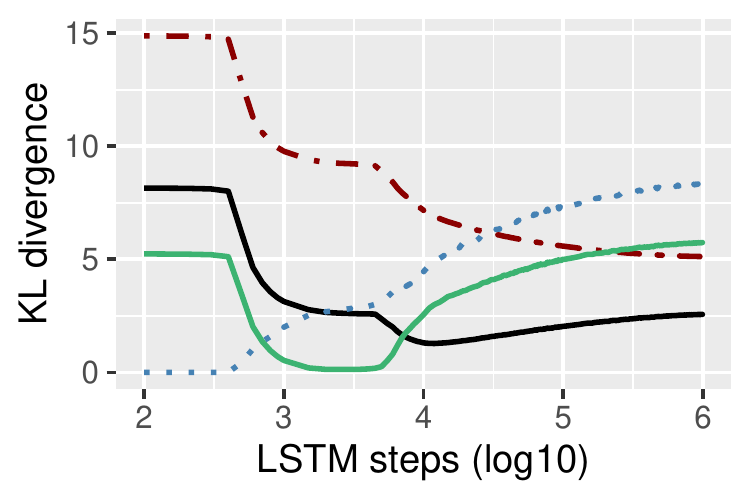}
    \includegraphics[width=5.7cm]{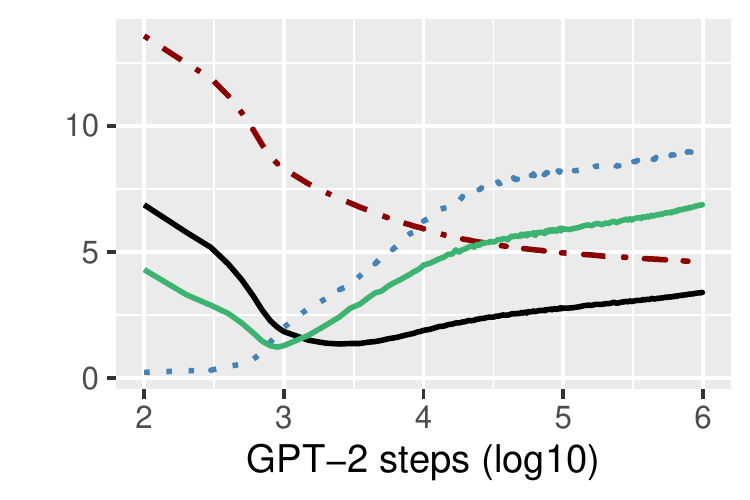}
    \includegraphics[width=3cm]{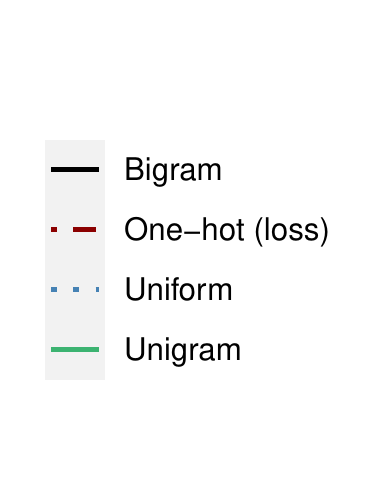} \newline
    \phantom{\_}\hspace{0.5cm}
    \includegraphics[width=5.7cm]{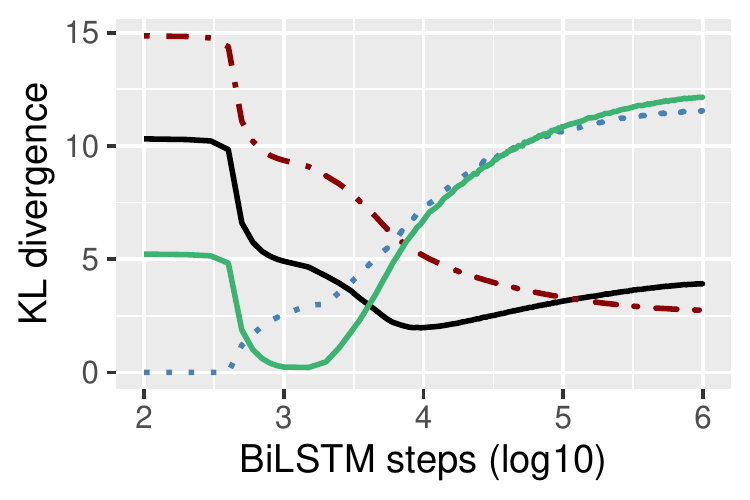}
    \includegraphics[width=5.7cm]{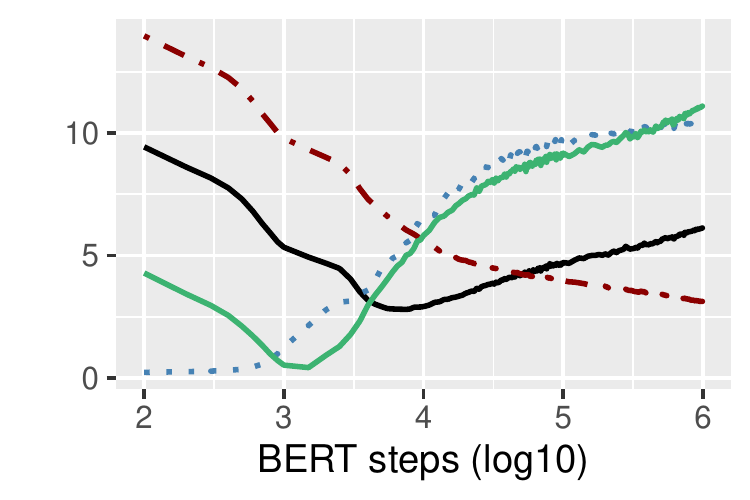}
    \caption{KL divergences between reference distributions and model predictions over the course of training.
    The KL divergence with the one-hot true token distribution is equivalent to the base two cross-entropy loss.
    Early in training, the models temporarily overfitted to unigram then bigram probabilities.}
    \label{fig:kl-curves}
\end{figure*}

As shown in Figure \ref{fig:kl-curves}, we plotted the KL divergences between each reference distribution and the model predictions over the course of training.
As expected, all four language models converged towards the true token distribution (minimizing the loss function) throughout training, diverging from the uniform distribution.
Divergence from the uniform distribution could also reflect that the models became more confident in their predictions during training, leading to lower entropy predictions.

As hypothesized, we also found that all four language models exhibited an early stage of training in which their predictions approached the unigram distribution, before diverging to reflect other information.
This suggests that the models overfitted to raw token frequencies early in training, an effect which was particularly pronounced in the LSTM-based models.
Importantly, because the models eventually diverged from the unigram distribution, the initial unigram  phase cannot be explained solely by mutual information between the true token distribution and unigram frequencies.

\subsection{Bigram probabilities}
We then ran a similar analysis using bigram probabilities, where each token probability was dependent only on the previous token.
A bigram distribution $P_b$ was computed for each masked token in the evaluation dataset, based on bigram counts in the training corpus.
As dictated by the bigram model definition, we defined $P_b(w_i) = P(w_i | w_{i-1})$ for unidirectional models, and $P_b(w_i) = P_b(w_i | w_{i-1}, w_{i+1}) \propto P(w_i | w_{i-1}) P(w_{i+1} | w_i)$ for bidirectional models.
We computed the average KL divergence between the bigram probability distributions and the language model predictions.

As shown in Figure \ref{fig:kl-curves}, during the unigram learning phase, the bigram KL divergence decreased for all language models.
This is likely caused by mutual information between the unigram and bigram distributions; as the models approached the unigram distribution, their divergences with the bigram distributions roughly approximated the average KL divergence between the bigram and unigram distributions themselves (average $\textrm{KL} = 3.86$ between unidirectional bigrams and unigrams; average $\textrm{KL} = 5.88$ between bidirectional bigrams and unigrams).
In other words, the models' initial decreases in bigram KL divergences can be explained predominantly by unigram frequency learning.

However, when the models began to diverge from the unigram distribution, they continued to approach the bigram distributions.
Each model then hit a local minimum in average bigram KL divergence before diverging from the bigram distributions.
This suggests that the models overfitted to bigram probabilities after the unigram learning phase.
Thus, it appears that early in training, language models make predictions based on unigram frequencies, then bigram probabilities, eventually learning to make more nuanced predictions.

Of course, this result may not be surprising for LSTM-based language models.
Because tokens are fed into LSTMs sequentially, it is intuitive that they would make use of bigram probabilities.
Our results confirm this intuition, and they further show that Transformer language models follow a similar pattern.
Because BERT and GPT-2 only encode token position information through learned absolute position embeddings before the first self-attention layer, they have no architectural reason to overfit to bigram probabilities based on adjacent tokens.\footnote{Absolute position embeddings in the Transformers were randomly initialized at the beginning of training.}
Instead, unigram and bigram learning may be a natural consequence of the language modeling task, or even distributional learning more generally.

\section{Discussion}
We found that language models are highly sensitive to basic statistics such as frequency and bigram probabilities during training.
Their acquisition of words is also sensitive to features such as sentence length and (for unidirectional models) lexical class.
Importantly, the language models exhibited notable differences with children in the effects of lexical class, word lengths, and concreteness, highlighting the importance of social, cognitive, and sensorimotor experience in child language development.

\subsection{Distributional learning, language modeling, and NLU}
In this section, we address the broader relationship between distributional language acquisition and contemporary language models.

\paragraph{Distributional learning in people}
There is ongoing work assessing distributional mechanisms in human language learning \cite{aslin-newport-2014-distributional}.
For instance, adults can learn syntactic categories using distributional information alone \cite{reeder-etal-2017-distributional}.
Adults also show effects of distributional probabilities in reading times \cite{goodkind-bicknell-2018-predictive} and neural responses \cite{frank-etal-2015-erp}.
In early language acquisition, there is evidence that children are sensitive to transition (bigram) probabilities between phonemes and between words \cite{romberg-saffran-2010-statistical}, but it remains an open question to what extent distributional mechanisms can explain effects of other factors (e.g. utterance lengths and lexical classes) known to influence naturalistic language learning.

To shed light on this question, we considered neural language models as distributional language learners.
If analogous distributional learning mechanisms were involved in children and language models, then we would observe similar word acquisition patterns in children and the models.
Our results demonstrate that a purely distributional learner would be far more reliant on frequency than children are.
Furthermore, while the effects of utterance length on words' ages of acquisition in children can potentially be explained by distributional mechanisms, the effects of word length, concreteness, and lexical class cannot.

\paragraph{Distributional models}
Studying language acquisition in distributional models also has implications for core NLP research.
Pre-trained language models trained only on text data have become central to state-of-the-art NLP systems.
Language models even outperform humans on some tasks \cite{he-etal-2020-deberta}, making it difficult to pinpoint why they perform poorly in other areas.
In this work, we isolated ways that language models differ from children in how they acquire words, emphasizing the importance of sensorimotor experience and cognitive development for human-like language acquisition.
Future work could investigate the acquisition of syntactic structures or semantic information in language models.

\paragraph{Non-distributional learning}
We showed that distributional language models acquire words in very different ways from children.
Notably, children's linguistic experience is grounded in sensorimotor and cognitive experience.
Children as young as ten months old learn word-object pairings, mapping novel words onto perceptually salient objects \cite{pruden-etal-2006-birth}.
By the age of two, they are able to integrate social cues such as eye gaze, pointing, and joint attention \cite{cetincelik-etal-2021-eyes}.
Neural network models of one-word child utterances exhibit vocabulary acquisition trajectories similar to children when only using features from conceptual categories and relations \cite{nyamapfene-ahmad-2007-multimodal}.
Our work shows that these grounded and interactive features impact child word acquisition in ways that cannot be explained solely by intra-linguistic signals.

That said, there is a growing body of work grounding language models using multimodal information and world knowledge.
Language models trained on visual and linguistic inputs have achieved state-of-the-art performance on visual question answering tasks (\citealp{antol-etal-2015-vqa}; \citealp{lu-etal-2019-vilbert}; \citealp{zellers-etal-2021-merlot}), and models equipped with physical dynamics modules are more accurate than standard language models at modeling world dynamics \cite{zellers-etal-2021-piglet}.
There has also been work building models directly for non-distributional tasks; reinforcement learning can be used for navigation and multi-agent communication tasks involving language (\citealp{chevalier-etal-2019-baby}; \citealp{lazaridou-etal-2017-multi}; \citealp{zhu-etal-2020-babywalk}).
These models highlight the grounded, interactive, and communicative nature of language.
Indeed, these non-distributional properties may be essential to more human-like natural language understanding (\citealp{bender-koller-2020-climbing}; \citealp{emerson-2020-goals}).
Based on our results for word acquisition in language models, it is possible that these multimodal and non-distributional models could also exhibit more human-like language acquisition.

\section{Conclusion}
In this work, we identified factors that predict words' ages of acquisition in contemporary language models.
We found contrasting effects of lexical class, word length, and concreteness in children and language models, and we observed much larger effects of frequency in the models than in children.
Furthermore, we identified ways that language models aquire unigram and bigram statistics early in training.
This work paves the way for future research integrating language acquisition and natural language understanding.

\section*{Acknowledgements}
We would like to thank the anonymous reviewers for their helpful suggestions, and the Language and Cognition Lab (Sean Trott, James Michaelov, and Cameron Jones) for valuable discussion.
We are also grateful to Zhuowen Tu and the Machine Learning, Perception, and Cognition Lab for computing resources.
Tyler Chang is partially supported by the UCSD HDSI graduate fellowship.

\bibliography{tacl2018,anthology,custom}
\bibliographystyle{acl_natbib}

\appendix
\section{Appendix}

\setlength\tabcolsep{3pt}
\begin{table}[t]
    \centering
    \small
    \renewcommand{\arraystretch}{1.11}
    \begin{tabular}{|>{\raggedright}p{5cm}|r|}
        \hline
        \textbf{Hyperparameter} & \textbf{Value} \\
        \hline
        Hidden size & 768 \\
        Embedding size & 768 \\
        Vocab size & 30004 \\
        Max sequence length & 128 \\
        Batch size & 128 \\
        Train steps & 1M \\
        Learning rate decay & Linear \\
        Warmup steps & 10000 \\
        Learning rate & 1e-4 \\
        Adam $\epsilon$ & 1e-6 \\
        Adam $\beta_1$ & 0.9 \\
        Adam $\beta_2$ & 0.999 \\
        Dropout & 0.1 \\ & \\
        \hline
        \textbf{Transformer hyperparameter} & \textbf{Value} \\
        \hline
        Transformer layers & 12 \\
        Intermediate hidden size & 3072 \\
        Attention heads & 12 \\
        Attention head size & 64 \\
        Attention dropout & 0.1 \\
        BERT mask proportion & 0.15 \\ & \\
        \hline
        \textbf{LSTM hyperparameter} & \textbf{Value} \\
        \hline
        LSTM layers & 3 \\
        Context size & 768 \\
        \hline
    \end{tabular}
    \caption{Language model training hyperparameters.}
    \label{tab:train-hyperparams}
\end{table}

\subsection{Language model training details}
\label{app:training-details}
Language model training hyperparameters are listed in Table \ref{tab:train-hyperparams}.
Input and output token embeddings were tied in all models.
Each model was trained using four Titan Xp GPUs.
The LSTM, BiLSTM, BERT, and GPT-2 models took four, five, seven, and eleven days to train respectively.

\setlength\tabcolsep{3pt}
\begin{table*}[!t]
    \centering
    \small
    \renewcommand{\arraystretch}{1.2}
    \begin{tabular}{|c|c|c|}
        \hline
        \textbf{LSTM} & \textbf{GPT-2} & \textbf{Children} \\
        \hline
        Adj $<$ Function $^{***}$ & Adj $<$ Function $^{***}$ & Nouns $<$ Adj $^{***}$ \\ 
        Adj $<$ Nouns $^{**}$ & Adj $<$ Other $^{*}$ & Nouns $<$ Verbs $^{***}$ \\
        Adj $<$ Other $^{**}$ & Verbs $<$ Function $^{***}$ & Nouns $<$ Function $^{***}$ \\
        Verbs $<$ Function $^{***}$ & Verbs $<$ Nouns $^{*}$ & Function $>$ Adj $^{***}$ \\
        Verbs $<$ Nouns $^{**}$ & Verbs $<$ Other $^{**}$ & Function $>$ Verbs $^{***}$ \\
        Verbs $<$ Other $^{*}$ & Nouns $<$ Function $^{***}$ & Function $>$ Other $^{***}$ \\
        & & Other $<$ Adj $^{**}$ \\
        & & Other $<$ Verbs $^{**}$ \\
        \hline
    \end{tabular}
    \caption{Significant pairwise differences between lexical classes when predicting words' ages of acquisition in language models and children (adjusted $p < 0.05^{*}$; $p < 0.01^{**}$; $p < 0.001^{***}$).
    A higher value indicates that a lexical class is acquired later on average.
    The five possible lexical classes were Noun, Verb, Adjective (Adj), Function Word (Function), and Other.
    }
    \label{tab:ancova}
\end{table*}

\begin{figure}[t]
    \includegraphics[width=7.5cm]{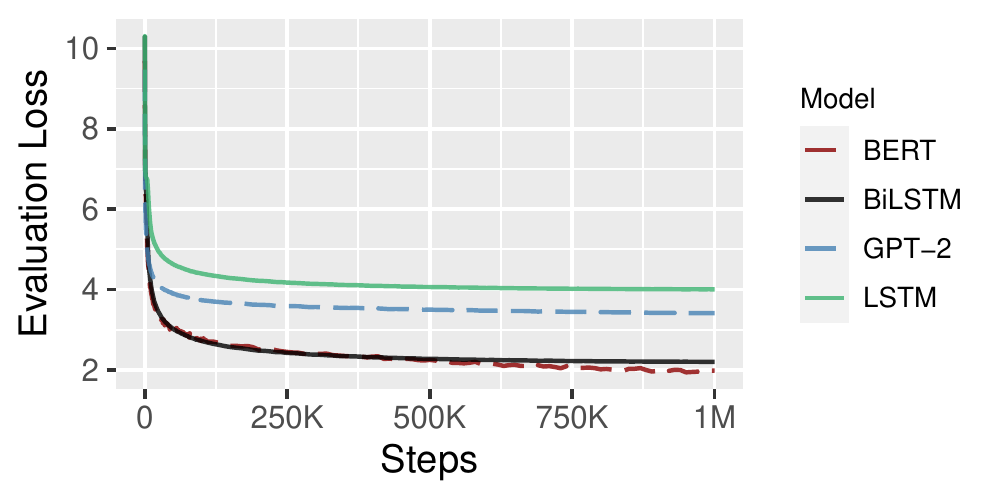}
    \caption{Evaluation loss during training for all four language models.
    Note that perplexity is equal to $\exp(\textit{loss})$.}
    \label{fig:loss-curves}
\end{figure}

To verify language model convergence, we plotted evaluation loss curves, as in Figure \ref{fig:loss-curves}.
To ensure that our language models reached performance levels comparable to contemporary language models, in Table \ref{tab:perplexity-comparisons} we report perplexity comparisons between our trained models and models with the same architectures in previous work.
For BERT, we evaluated the perplexity of Huggingface's pre-trained BERT base uncased model on our evaluation dataset \cite{wolf-etal-2020-transformers}.
For the remaining models, we used the evaluation perplexities reported in the original papers: \citet{gulordava-etal-2018-colorless} for the LSTM,\footnote{The large parameter count for the LSTM in \citet{gulordava-etal-2018-colorless} is primarily due to its large vocabulary without a decreased embedding size.} \citet{radford-2019-language} for GPT-2 (using the comparably-sized model evaluated on the WikiText-103 dataset), and \citet{aina-etal-2019-putting} for the BiLSTM.
Because these last three models were cased, we could not evaluate them directly on our uncased evaluation set.
Due to differing vocabularies, hyperparameters, and datasets, our perplexity comparisons are not definitive; however, they show that our models perform similarly to contemporary language models.

Finally, we evaluated each of our models for word acquisition at 208 checkpoint steps during training, sampling more heavily from earlier steps.
We evaluated checkpoints at the following steps:
\begin{itemize}[noitemsep]
    \item Every 100 steps during the first 1000 steps.
    \item Every 500 steps during the first 10,000 steps.
    \item Every 1000 steps during the first 100,000 steps.
    \item Every 10,000 steps for the remainder of training (ending at 1M steps).
\end{itemize}

\begin{table}[t]
    \centering
    \footnotesize
    \renewcommand{\arraystretch}{1.1}
    \begin{tabular}{|r|c|c|c|c|}
        \cline{2-5}
        \multicolumn{1}{c|}{} & \multicolumn{2}{c|}{Ours} &
        \multicolumn{2}{c|}{Previous work} \\
        \cline{2-5}
        \multicolumn{1}{c|}{} & \# Params & Perplexity & \# Params & Perplexity \\
        \hline
        LSTM & 37M & 54.8 & 72M $^a$ & 52.1 \\
        GPT-2 & 108M & 30.2 & 117M $^b$ & 37.5 \\
        BiLSTM & 51M & 9.0 & 42M $^c$ & 18.1 \\
        BERT & 109M & 7.2 & 110M $^d$ & 9.4 \\
        \hline
    \end{tabular}
    \caption{Rough perplexity comparisons between our trained language models and models with the same architectures in previous work ($^a$\citealp{gulordava-etal-2018-colorless}; $^b$\citealp{radford-2019-language}; $^c$\citealp{aina-etal-2019-putting}; $^d$\citealp{wolf-etal-2020-transformers}).
    }
    \label{tab:perplexity-comparisons}
\end{table}

\subsection{Lexical class comparisons}
\label{app:pairwise}
We assessed the effect of lexical class on age of acquisition in children and each language model.
As described in the text, when lexical class reached significance based on the likelihood ratio test (accounting for log-frequency, MLU, n-chars, and concreteness), we ran a one-way analysis of covariance (ANCOVA) with log-frequency as a covariate.
There was a significant effect of lexical class in children and the unidirectional language models (the LSTM and GPT-2; $p < 0.001$).

Pairwise differences between lexical classes were assessed using Tukey's honestly significant difference (HSD) test.
Significant pairwise differences are listed in Table \ref{tab:ancova}.

\end{document}